\pdfoutput=1
\documentclass[lettersize,journal]{IEEEtran}
\usepackage{amsmath,amsfonts}
\usepackage[ruled,linesnumbered]{algorithm2e}
\usepackage{array}
\usepackage[caption=false,font=normalsize,labelfont=sf,textfont=sf]{subfig}
\usepackage{textcomp}
\usepackage{stfloats}
\usepackage{bm}
\usepackage{url}
\usepackage{verbatim}
\usepackage{graphicx}
\usepackage{cite}
\usepackage{multirow}
\usepackage[table,xcdraw]{xcolor}
\usepackage{threeparttable}
\usepackage[implicit=false]{hyperref}
\usepackage{hyperref}
\hypersetup{hidelinks,
	colorlinks=true,
	allcolors=black,
	pdfstartview=Fit,
	breaklinks=true}

\newcommand{\orcid}[1]{\href{https://orcid.org/#1}{\includegraphics[width=10pt]{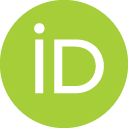}}}

\newcommand*{\minorchange}{\textcolor{black}}
\newcommand*{\bigchange}{\textcolor{black}}

\makeatletter
\def\bstctlcite{\@ifnextchar[{\@bstctlcite}{\@bstctlcite[@auxout]}}
\def\@bstctlcite[#1]#2{\@bsphack
  \@for\@citeb:=#2\do{%
    \edef\@citeb{\expandafter\@firstofone\@citeb}%
    \if@filesw\immediate\write\csname #1\endcsname{\string\citation{\@citeb}}\fi}%
  \@esphack}
\makeatother

\begin{document}

\bstctlcite{BSTcontrol}

\title{FireFly: A High-Throughput Hardware Accelerator for Spiking Neural Networks with Efficient DSP and Memory Optimization
%FireFly: A DSP-Optimized FPGA Accelerator for Spiking Neural Networks with Efficient Memory Access
}

\author{
    Jindong Li \orcid{0000-0002-4009-916X},
    Guobin Shen \orcid{0000-0002-4069-2107},
    Dongcheng Zhao \orcid{0000-0002-0593-8650}, 
    Qian Zhang \orcid{0000-0001-5314-4233},
    Yi Zeng \orcid{0000-0002-9595-9091}

\thanks{Manuscript created 1 January 2023; revised 16 April 2023; accepted 16 May 2023. This work was supported by the Strategic Priority Research Program of the Chinese Academy of Sciences (XDB32070100); the Chinese Academy of Sciences Foundation Frontier Scientific Research Program (ZDBS-LY- JSC013). \textit{(Corresponding authors: Qian Zhang; Yi Zeng.)}}

\thanks{Jindong Li and Qian Zhang are with the School of Artificial Intelligence, University of Chinese Academy of Sciences, Beijing 100049, China, and also with the Brain-inspired Cognitive Intelligence Lab, Institute of Automation, Chinese Academy of Sciences, Beijing 100190, China (e-mail: lijindong2022@ia.ac.cn, q.zhang@ia.ac.cn).}

\thanks{Guobin Shen is with the School of Future Technology, University of Chinese Academy of Sciences, Beijing 100049, China, and also with the Brain-inspired Cognitive Intelligence Lab,  Institute of Automation, Chinese Academy of Sciences, Beijing 100190, China (e-mail: shenguobin2021@ia.ac.cn).}

\thanks{Dongcheng Zhao is with the Brain-inspired Cognitive Intelligence Lab, Institute of Automation, Chinese Academy of Sciences, Beijing 100190, China (e-mail: zhaodongcheng2016@ia.ac.cn).}  

\thanks{Yi Zeng is with the Brain-inspired Cognitive Intelligence Lab, Institute of Automation, Chinese Academy of Sciences, Beijing 100190, China, and University of Chinese Academy of Sciences, Beijing 100049, China, and Center for Excellence in Brain Science and Intelligence Technology, Chinese Academy of Sciences, Shanghai 200031, China (e-mail: yi.zeng@ia.ac.cn).}
}

\maketitle

\begin{abstract}
Spiking neural networks (SNNs) have been widely used due to their strong biological interpretability and high energy efficiency. With the introduction of the backpropagation algorithm and surrogate gradient, the structure of spiking neural networks has become more complex, and the performance gap with artificial neural networks has gradually decreased. 
However, most SNN hardware implementations for field-programmable gate arrays (FPGAs) cannot meet arithmetic or memory efficiency requirements, which significantly restricts the development of SNNs. They do not delve into the arithmetic operations between the binary spikes and synaptic weights or assume unlimited on-chip RAM resources by using overly expensive devices on small tasks.
To improve arithmetic efficiency, we analyze the neural dynamics of spiking neurons, generalize the SNN arithmetic operation to the multiplex-accumulate operation, and propose a high-performance implementation of such operation by utilizing the DSP48E2 hard block in Xilinx Ultrascale FPGAs. 
To improve memory efficiency, we design a memory system to enable efficient synaptic weights and membrane voltage memory access with reasonable on-chip RAM consumption. Combining the above two improvements, we propose an FPGA accelerator that can process spikes generated by the firing neurons on-the-fly (FireFly).
\minorchange{FireFly is the first SNN accelerator that incorporates DSP optimization techniques into SNN synaptic operations.}
FireFly is implemented on several FPGA edge devices with limited resources but still guarantees a peak performance of 5.53 TOP/s at 300MHz. As a lightweight accelerator, FireFly achieves the highest computational density efficiency compared with existing research using large FPGA devices.

%To the best of our knowledge, FireFly achieves the highest throughput compared with the existing research.
%achieving a balanced resource consumption between LUTs and DSPs.
\end{abstract}


\section{Introduction}
\IEEEPARstart{S}{piking} neural networks (SNNs) are considered as the third generation of artificial neural networks (ANNs)~\cite{maass1997networks}. They were developed to mimic the operational mechanism in the human brain, where information is communicated via spikes among neurons.
%Surrogate gradient algorithms have been introduced for SNNs tackling nondifferentiable problems to enhance the learning capability of SNNs.~\cite{wu2018spatio,zhang2020temporal}.
Recent advances in SNNs have demonstrated comparable performance to non-spiking ANNs~\cite{shen2022backpropagation,zheng2021going}. However, compared to the extensive work on ANN accelerators~\cite{chen2016eyeriss,samajdar2019scaling,wu2017high}, the existing SNN hardware accelerator still lags, limiting the practical applications of SNNs.% In this paper, we focus on two critical and often overlooked aspects of SNN accelerator design targeting FPGAs.

Most research ignores the importance of efficiently implementing arithmetic operations in SNN accelerators. In Field-programmable gate array (FPGA) design, using the built-in dedicated hard block to implement arithmetic operations can achieve considerably higher performance than its general logic fabric counterparts. Fabric-only implementations in an arithmetic-extensive application can lead to a compromised clock frequency and even routing failures when the fabric consumption is high. However, in the SNN accelerator design, the register transfer level (RTL) description of the SNN arithmetic operation cannot be automatically synthesized into the dedicated arithmetic hard block. Therefore, most SNN accelerators adopt the fabric-only implementation without further optimizations. Although a single arithmetic operation unit in an SNN accelerator consumes considerably fewer resources than a multiply-accumulate (MAC) unit in an ANN accelerator design, hardware optimization of such operation can still significantly impact the system's performance when the unit is instantiated hundreds or even thousands of times. In the Xilinx Ultrascale FPGA, the dedicated arithmetic hard block, or the DSP48E2, enhances the speed and efficiency of many operations, including multiplication, addition, wide bus multiplexing, pattern detection, and single instruction multiple data (SIMD) operations. It is possible to generalize the SNN computation to the arithmetic operations that the DSP48E2 can provide.

Another important aspect of the SNN accelerator design is the memory system. When scaling the parallelism, the memory bandwidth imbalance between the binary input-output spikes, the multi-bit synaptic weights and the multi-bit membrane voltage becomes problematic. While the computational complexity and the memory footprint of the binary spikes decrease, the memory access requirements of synaptic weights and membrane voltage do not. The off-chip memory access bandwidth needed by the weights and membrane voltage cannot fully support the increased parallelism brought by the hardware-friendly synaptic operations and storage-friendly binary spikes without further exploration of the reuse mechanism. Most hardware accelerators assume large on-chip memory, store all the synaptic weights, and accumulate membrane voltage on-chip to ease the harsh bandwidth requirement. This method is not scalable, especially when the model gets larger and targets edge FPGA devices. A scalable memory system for synaptic weights and membrane voltage balancing, as well as off-chip data access and on-chip data buffering, should be developed.

% At present, most existing neuromorphic hardware or accelerators focus on brain simulation tasks. While these hardware designs claim to support event-driven processing, they are inefficient in terms of resource utilization, computational density, and scalability. In real-world SNN applications, it is not feasible to use overly expensive and large FPGA devices. A lightweight and high-performance SNN accelerator targeting resource-constrained edge scenarios should be developed.

At present, most existing neuromorphic hardware or accelerators are inefficient in terms of resource utilization, computational density, and scalability. In real-world SNN applications, it is not feasible to use overly expensive and large FPGA devices. A lightweight and high-performance SNN accelerator targeting resource-constrained edge scenarios should be developed.
Focusing on these aspects, we propose FireFly, a high throughput and reconfigurable FPGA accelerator that can process spikes generated by the \underline{firing} neurons on-the-\underline{fly}, achieving both arithmetic and memory efficiency. Our contributions can be summarized as follows.

\begin{enumerate}
\item{
We generalize the SNN arithmetic operation to the multiplex-accumulate operation and propose a high-performance implementation of such an operation by utilizing the DSP48E2 hard block in Xilinx Ultrascale FPGAs.
}
\item{
We design a synaptic weight delivery hierarchy and a partial sum and membrane voltage (Psum-Vmem) unified buffer to balance the off-chip memory access bandwidth and on-chip RAM consumption.
}
\item{
We evaluate multiple deep SNN models on various datasets and achieve faster inference speed and higher classification accuracy than the existing research. We implement FireFly on several commercial off-the-shelf FPGA edge devices with limited resources, bringing hope for real-world SNN applications in edge scenarios. 
}
\end{enumerate}

\section{Related Work}
The existing dedicated neuromorphic hardware designed for SNN can be categorized into four types.

The majority of neuromorphic hardware constructs its hardware substrates in a Network on Chip fashion. Spinnaker\cite{painkras2013spinnaker}, Loihi\cite{davies2018loihi}, and TrueNorth\cite{akopyan2015truenorth} fall into this category. In these hardware designs, neurons are grouped into multiple neurocores, which communicate via spikes through the Network-on-Chip (NoC), and spike messages are scheduled by dedicated routers. These hardware architectures are compatible with the event-driven nature of SNNs, as spike events are generated, transferred, and processed only if the neuron fires. However, these neuromorphic hardware designs place rigid restrictions on the network. The SNN networks are distributed among the neurocores, and the total number of neurons in the model cannot exceed the maximum capacity of the hardware, not to mention the harsh fan-in and fan-out hardware limitations of the network.

The second type of neuromorphic hardware explores emerging devices. The BrainScale\cite{schemmel2010wafer} developed by Heidelberg University emulated spiking neural networks on analog neuromorphic hardware and achieved several advantages over conventional computers. Some research explores new materials like mem-resistors and optics \cite{feldmann2019all,yang2020leaky}. However, the low precision and uncertain nature of the hardware prevent them from being used in practice.

The third type of neuromorphic hardware follows the scheme of the ANN accelerator design except for constructing dedicated hardware for synaptic operations and explores optimal dataflow for SNNs specifically\cite{chen2022cerebron,ye2022implementation,narayanan2020spinalflow,liu2022sato}. These types of work require less area cost and achieve higher computing resource utilization. Fine-grained parallelism of the accelerator can enable high-performance computing of the SNN compared with the sequential spike processing mechanism of the NoC counterparts. This type of hardware has the fewest restrictions on the network models and can quickly adapt to emerging neuromorphic research.
FPGA platforms are the ideal choice for this type of hardware due to their flexibility and reconfigurability.

\minorchange{
The three types of neuromorphic hardware designs listed above have a general hardware architecture that can adapt to different types of networks, whereas the fourth type of neuromorphic hardware is tailored to particular networks \cite{panchapakesan2022syncnn,aung2021deepfire,park201965,chuang202090nm}. Park et al.\cite{park201965} build an on-chip learning system tailored for a two-layer SNN using direct spike-only feedback. Chuang et al.\cite{chuang202090nm} introduce a low-power 90nm CMOS binary weight spiking neural network ASIC for real-time image classification.
\cite{panchapakesan2022syncnn,aung2021deepfire} target FPGA devices and design inference engines for specific neural networks.
Although these hardware designs can achieve high energy efficiency and inference speed, they are limited in their practicality for deep and large SNNs due to their linear expansion in power and area as network size increases. Moreover, because ASIC designs lack reconfigurability, hardware specifically designed for one network may not be adaptable to other network configurations.
}

While FireFly belongs to the third category, FireFly's contributions are largely complementary to the existing work.
SyncNN~\cite{panchapakesan2022syncnn} proposed a novel synchronous event-driven SNN reconfigurable inference engine and evaluated multiple SNN models on multiple FPGA devices. Fang et al. \cite{fang2020encoding} proposed a holistic optimization framework for the encoder, model, and architecture design of FPGA-based neuromorphic hardware. However, these designs are based on high-level synthesis, thus inducing large resource redundancy.
Lee et al.~\cite{lee2022parallel} and Chen et al. ~\cite{chen2022skydiver} explored spatial-temporal parallelism by unrolling the computations in both the spatial and time dimensions and achieved significant acceleration. However, parallelization across multiple time points violates the time-related sequential nature of the membrane voltage update behavior.
SpinalFlow~\cite{narayanan2020spinalflow} achieved significant sparsity acceleration by adopting a different input/output spike representation to skip the non-spike computations. SATO~\cite{liu2022sato} achieved high-speed inference by incorporating a temporal-oriented dataflow and a bucket-sort-based dispatcher to balance the workload. However, these techniques only work for temporal coding SNNs, limiting the accuracy of the SNN models.
DeepFire~\cite{aung2021deepfire} was the first research migrating DSP48E2s into neuron core design. However, they did not delve into the function of DSP48E2 and still induce large fabric overhead.

We argue that with careful register transfer level (RTL) design, focusing on optimizing spatial parallelism on FPGA, adopting regular and simple time-step CNN-like processing, and fully utilizing the multi-function DSP48E2, we can still achieve impressive inference throughput on small FPGA edge devices. FireFly is more applicable in real-world applications where design space exploration is constrained by limited resources.

\section{SNN Basics}

\subsection{Spiking Neuron Model}
Spiking neurons are the basic units of SNNs, which are connected through weighted synapses and transmit information through binary spikes. Although more complex and detailed neuron models such as Izhikevich\cite{izhikevich2004model} and Hodgkin–Huxley\cite{hodgkin1952quantitative} can accurately model a biological neuron’s behavior, simpler models such as Integrate and Fire (IF)\cite{abbott1999lapicque} and Leaky Integrate and Fire (LIF)\cite{dayan2003theoretical} are used more often in current SNN applications.

An IF neuron integrates its inputs over multiple timesteps and generates a spike whenever the integrated membrane voltage surpasses a firing threshold. A LIF neuron acts the same except for the leaky behavior of the membrane voltage. 
The neural dynamics of a LIF neuron membrane potential $u$ can be described as:

\begin{equation}
    \tau_m \frac{d u}{d t}=-u+R \cdot I(t), \quad u<V_{th}
    \end{equation}

where $V_{th}$ denotes the threshold, $I$ denotes the input current,  $R$ denotes the resistance, and $\tau_m$ is the membrane time constant. A spike is generated when $u$ reaches $V_{th}$ and $u$ is reset to resting potential $u_{rest}$, which is set to 0 in this work. The membrane potential's neural dynamics can be divided into three phases, and each phase can be described in a discrete computational form::

\emph{Input current integration phase.} All the presynaptic currents generated by the presynaptic spikes are integrated at each discrete timestep.
\begin{equation}
    I[t]=\sum_j w_{i j} s_j[t]+b_i
    \end{equation}
where the subscript $i$ represents the
$i_{th}$ neuron, $w_{ij}$ is the synaptic weight from neuron $j$ to neuron $i$, and $b_i$ is a bias.

\emph{Membrane potential update phase.} The membrane potential of each neuron is updated by the integrated presynaptic currents at each timestep. 
\begin{equation}
    v_i[t]=(1-\frac{1}{\tau_m}) u_i[t] +I[t]
    \end{equation}
where $(1-\frac{1}{\tau_m}) < 1$ denotes the leaky term, which is ignored when using the IF model.

\emph{Output spike generation phase.} Whenever the membrane potential reaches the firing threshold, the neuron generates an output spike and resets its membrane potential.
\begin{gather}\label{lif}
    (u_i[t+1],s_i[t+1])= \left\{\begin{array}{l}
        (v_i[t],0),v_i[t] < V_{t h}\\
        (0,1), \quad \  v_i[t] \geq V_{t h}
    \end{array}\right.
\end{gather}

In these three phases, we have two key observations. The input current integration phase completely dominates the total computational cost due to the high degree of synaptic connectivity and a large number of neurons. The membrane potential update phase has the harshest storage requirement because the membrane potential is read and written back and forth in every timestep. We will focus on these two aspects in the following sections.

\subsection{Dataflow and Parallelism Scheme for SCNN}

Similar to convolutional neural networks (CNNs), convolutional layers dominate the total computational cost in spiking convolutional neural networks (SCNNs). We mainly focus on the dataflow optimizations of the convolutional layers and show that the dataflow can be migrated to fully connected layers.

Input/Output spike representation varies in different neuromorphic hardware. Most SNN hardware implementations adopt the Address-Event-Representation (AER) data format to transmit spikes between neurons. The standard AER package for one spike includes the spiking neuron's input location and the spike's timestamp. Although the AER data format is compatible with the event-driven nature of SNNs, multiple bits are needed to express the original single-bit spike event. The logic and storage overhead may not be worth it.

This paper adopts the original single-bit format to represent the binary spikes. At any discrete timestep $t$ in the digitalized SCNN, the output spikes of all the neurons in one channel of the convolutional layer can be considered a timestep snapshot in the form of a binary map\cite{zhang2021cost}. In this case, the input-current integration phase computation process of the SNNs is almost the same as that of the traditional ANNs except for the additional time dimension and the changed operation. The set of computations for the complete SNN convolutional layer that receives a single batch of input can be formulated as a loop nest over these $7$ variables. All permutations of these $6$ loop variables, except for the timestep variable, are legal. Permutations of the loop variables open up the possibility of different dataflow choices. The tiling of the loop variables opens up the possibility of different parallelism schemes.

Different permutations of the loop variables adopt different kinds of dataflow. Different dataflow schemes for convolution have been extensively studied by Eyeriss\cite{chen2016eyeriss}. The key consideration is how to minimize data movement and maximize data reuse. In SCNN, synaptic connection weights need to be fetched and membrane voltage needs to be updated at every time timestep, due to the unique time dimension in SNN computation. Therefore, output and weight stationary (OS and WS) dataflow can minimize the data movement of the multi-bit membrane voltage and synaptic weight data between on-chip logic and off-chip memory.

Different tiling strategies for the loop variables enable different parallelism schemes. The tiling of the loop variables can induce data reordering or data segmentation. We argue that it is important to keep the input and output spike arrangements the same to enable spikes to be processed in an on-the-fly fashion without complicated data rearrangement. We chose the spatial tiling of the input and output channel dimensions rather than tiling within the same spike feature map to avoid data rearranging or irregular off-chip data access.

\begin{algorithm}

    \KwIn{ Given the binary spike map size $(H,W)$, input-output channels $(C_{in},C_{out})$, kernel size $(K_h,K_w)$, total timestep $T$, leaky factor $\lambda $, threshold $V_{th}$ and parallelism factor  $P$. Divide the input output channels into $(c_{i}=\lceil \frac{C_{in}}{P}\rceil ,c_{o}=\lceil \frac{C_{out}}{P}\rceil )$ groups.}
    
    \KwIn{$T\times c_i$ fragments of $I[H\times W][P]$ streams, each stream passes the hardware for $c_o$ times.}
    
    \KwOut{$T\times c_o$ fragments of $O[H\times W][P]$ streams.}
    Allocate buffer for synaptic weights: $W[P][C_{in}][K_h][K_w]$\;
    Allocate buffer for Psum/Vmem: $V[H\times W][P]$\;
    \For{$p_o \leftarrow 0$ \KwTo $c_o$}{
        Load Weights: $W[P][C_{in}][K_h][K_w]$\;
        \For{$t \leftarrow 0 $ \KwTo $T$}{
            \For{$p_i \leftarrow 0$ \KwTo $c_i$}{
                \For{$s \leftarrow 0$ \KwTo $L=H\times W$}{
                    \emph{Unroll and pipeline}\;
                    \For{$o \leftarrow 0$ \KwTo $P$}{
                        \For{$i \leftarrow 0$ \KwTo $P$}{
                            $\bm{w}=W[o][p_i\times P+i][0\rightarrow K_h][0\rightarrow K_w]$\;
                            $\bm{i}=neighbour\left(I[s][i]\right)$\;
                            $V[s][o]+=\bm{w}\cdot \bm{i}$\;
                        }
                    }
                    \uIf{$p_i=c_i -1$}{
                        \For{$o \leftarrow 0$ \KwTo $P$}{
                        $V[s][o]\times=\left(1-\lambda\right)$\;
                        \eIf{$V[s][o]>V_{th}$}{
                            $V[s][o]=0,O[s][o]=1$\;
                        }{
                            $O[s][o]=0$\;
                        }
                        \uIf{$t=T -1$}{
                            $V[s][o]=0$\;
                        }
                        }
                    }
                }
            }
        }
    }
    \caption{Pseudo Code of Scheduling a Single Convolutional Layer in FireFly Architecture.}
    \label{algo}
    \end{algorithm}

Adopting the dataflow and parallelism scheme above, the pseudo-code of scheduling a single convolutional layer in FireFly architecture is described in Algorithm \ref{algo}.
\minorchange{
Given an input spike tensor in shape $T\times C_{in} \times H \times W$, a weight tensor in shape $C_{out}\times C_{in}\times K_{h}\times K_{w}$ and an output spike tensor in shape $T\times C_{out} \times H \times W$ assuming same padding and stride of one, where $T$ denotes the timestep, $(C_{out}, C_{in})$ denotes the output-input channels, $(H, W)$ denotes the size of spike maps.
We adopt channel tiling in output and input channels with the parallelism factor $P$ and flatten the spike map to one dimension data stream of length $L=H\times W$, yielding $T\times c_i$ fragments of input spike stream with $P$ spike channels and $T\times c_o$ fragments of output spike stream with $P$ spike channels, where $c_i = \lceil \frac{C_{in}}{P}\rceil$ and $c_o = \lceil \frac{C_{out}}{P}\rceil$.
FireFly receives $P$ channels of input spike stream, performs a $P\times P$ spike map convolution and generates $P$ channels of output partial sum. The $P\times P$ convolutions are unrolled spatially, while $T$, $c_o$ and $c_i$ are folded in time, reusing the same hardware substrates.
Any permutation of the loop variables $T$, $c_o$ and $c_i$ is legal. We iterate $c_o$ over $T$ over $c_i$, adopting a weight stationary and output stationary dataflow as discussed above.
In this way, $c_i$ fragments of input spike stream pass the hardware for $c_o$ times at each timestep. The calculation of IF/LIF neural dynamics is performed when the last fragment of the input spike stream flows through. The membrane voltage is cleared when the integration and spike generation process is done for all timesteps.
}
\section{Hardware Architecture}
\begin{figure*}
    \centering
    \includegraphics[width=1.0\linewidth]{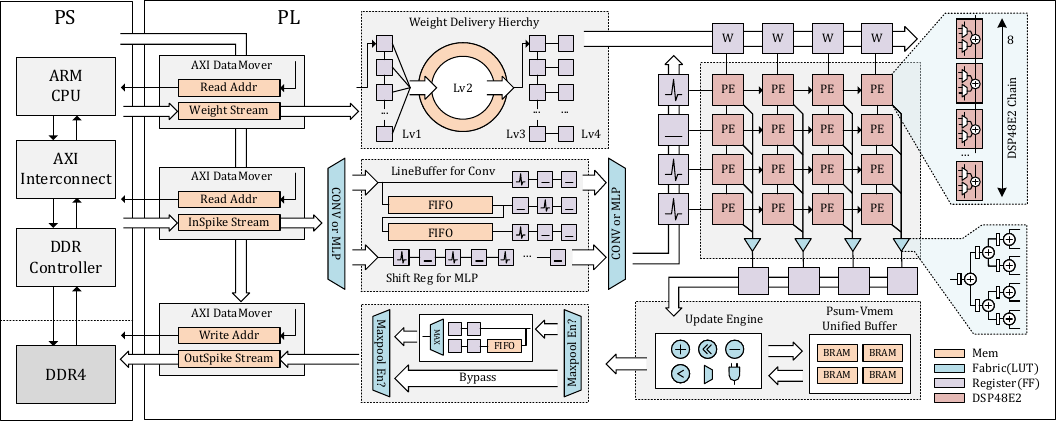}
    \caption{The Architecture of FireFly.}
    \label{fig:arch}
\end{figure*}

\subsection{Architecture Overview}
In this section, the digital design of SNNs is discussed in detail. Fig.\ref{fig:arch} shows the overall system design of FireFly.
FireFly targets heterogeneous Zynq Ultrascale devices. The central processing unit (CPU) of the processing system (PS) acts as the controller for system state control and external memory access. The programmable logic (PL) accelerates the SNN inference.
AXI DataMover IP, instead of AXI DMA IP, enables high-throughput and low-latency data transactions between the off-chip DRAM and on-chip memory storage.
The unique store and forward feature of AXI DataMover is enabled to allow multiple outstanding requests.

The weight-stationary systolic array is responsible for the acceleration of SNN arithmetic operations. The systolic array consists of several DSP48E2 chains and multiple adder trees. A weight matrix delivery hierarchy is proposed to enable efficient weight loading to the systolic array. Two separate datapaths for convolutional and fully connected layers are designed to generate binary spike vectors for the systolic array. A Psum-Vmem unified buffer and update engine is constructed to support back-and-forth membrane potential update and IF/LIF neuron dynamics. An optional MaxPooling unit is placed on the output spike datapath to support on-the-fly pooling.

The designs of the systolic array, the spike vector generation unit, the synaptic weight delivery hierarchy, and the Psum-Vmem unified buffer are elaborated in detail below.

\subsection{\bigchange{Synaptic Crossbar Computation Featured by DSP48E2s}}

\begin{figure*}
    \centering
    \includegraphics[width=1.0\linewidth]{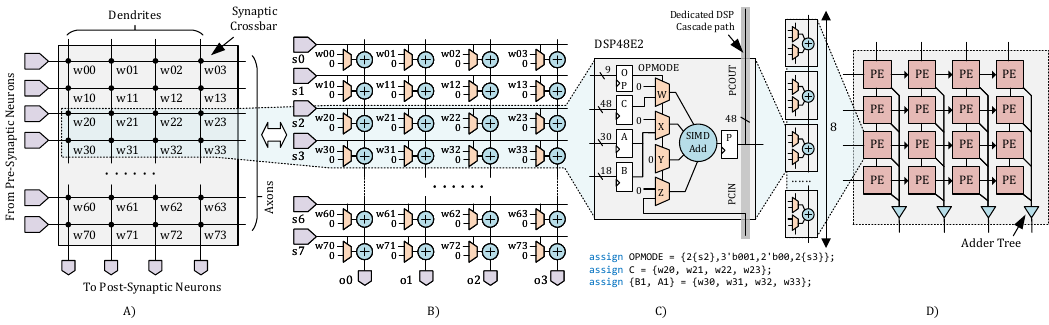}
    \caption{Synaptic Crossbar Computation. \minorchange{A) A $8 \times 8$ synaptic crossbar. B) The equivalent digital circuit of the $8 \times 8$ synaptic crossbar.} C) Implementation of a $2\times 4$ synaptic crossbar by a single DSP48E2 slice. \minorchange{D) A systolic array with $4 \times 4$ PEs. A PE is a DSP48E2 cascaded chain of length 8.}}
    \label{fig:dsp}
\end{figure*}

\minorchange{Fig.\ref{fig:dsp}A shows an all-to-all, fully-connected connection topology between 8 pre-synaptic neurons and 8 post-synaptic neurons. The axons of pre-synaptic neurons and the dendrites of the post-synaptic neurons are crossed, forming an $8 \times 8$ synapse matrix. The input spikes from the axon of a pre-synaptic neuron are broadcasted through a row of the crossbar. The dendrite integrates the input spikes along the column of the crossbar.}

\minorchange{The synaptic operation happens at every crossing point of the synaptic crossbar, represented by a black dot. Mathematically, the synaptic operation consists of a dot product between the binary spike and the synaptic weight and an addition accumulating the synaptic current propagated along the dendrites. Such an operation can be implemented by a multiplexer and an adder. The spike acts as the control signal of the multiplexer, switching the synaptic weight on or off when the neuron is firing or resting. The adder sums up the result from the multiplexer and the result coming from the cross point above. Fig.\ref{fig:dsp}B shows the equivalent digital circuit to the connection topology shown in Fig.\ref{fig:dsp}A.}
\minorchange{In ASIC design, the RTL description of the synaptic operation is synthesized into standard cells of a certain technology library. In FPGA design, the RTL description of the synaptic operation is automatically synthesized into LUTs and FFs. However, we show that the multiplex-accumulating operation existing in the crossbar computation can be manually mapped to the dedicated DSP hard block in Xilinx FPGA, leading to significant improvements in resource efficiency and clock rate.}

DSP48E2 is the dedicated digital signal processing logic block in the Xilinx Ultrascale series FPGA. Most FPGA neuromorphic hardware simply treats them as multipliers and leaves them underutilized. However, they enhance the speed and efficiency of many applications far beyond multiplication-based digital signal processing\cite{xilinx2021ultrascale}. In this chapter, we show that a single DSP48E2 slice can support a $2 \times 4$ synaptic crossbar computation, and up to 8 DSP48E2 slices can be cascaded in a chain to support a $16 \times 4$ crossbar computation without numeric overflow. By instantiating multiple DSP48E2 cascaded chains and arranging them in a 2D matrix, we can construct a systolic array supporting larger synaptic crossbar computation. The detailed implementation of this method is demonstrated below.

The DSP48E2 slice consists of four pipeline stages for input ports A, B, C and D, a 27-bit pre-adder, a $27 \times 18$ multiplier, four 48-bit wide-bus multiplexers named W, X, Y and Z and a flexible 48-bit ALU. A 5-bit INMODE port sets the configuration of the input pipeline stages and the 27-bit pre-adder. A 9-bit OPMODE port controls the select signal of the W, Y, X and Z multiplexer. A 4-bit ALUMODE port controls the functionality of the 48-bit ALU. In FireFly, we fully utilize the four 48-bit wide-bus multiplexers, dynamic control of the OPMODE and the SIMD mode of the 48-bit ALU to implement the crossbar computation.

The static configuration of the DSP48E2 is configured as below:
The 27-bit pre-adder and the $27 \times 18$ multiplier are disabled. 
The 4-bit ALUMODE port is set to 4'b0000 so that the ALU unit will perform add operation.
The 5-bit INMODE port is set to 5'b10001 so that data ports A and B are registered once. Data port C is registered once. Data port D is left unused.
All the carry inputs are ignored. 
The 48-bit ALU unit is configured into SIMD mode, supporting four independent 12-bit additions.
Direct access to these specific configurations in DSP48 is achieved by directly instantiating the DSP48E2 primitive.
In this way, the outputs of the four 48-bit multiplexers W, X, Y and Z are split into four 12-bit fields respectively. The 48-bit ALU unit acts as four independent 12-bit adders summing up each field of the four multiplexers.

The dynamic configuration of the DSP48E2 involves changing the OPMODE at runtime to switch the multiplexers to different inputs.
There are dozens of combinations of inputs to these multiplexers, one of them can be: either C or all 0s on the W multiplexer; either A:B or all 0s on the X multiplexer; all 0s on the Y multiplexer; PCIN on the Z multiplexer, where PCIN is the output of a lower DSP slice, cascaded into the current DSP slice.

\minorchange{Adopting the static and dynamic configuration of the DSP48E2 described above, the synaptic crossbar computation can be efficiently implemented by 3 levels of DSP48E2 instantiation: a single DSP48E2 slice, a DSP48E2 chain and a DSP48E2 systolic array. }

\subsubsection{\bigchange{Synaptic Crossbar Computation by a Single DSP48E2}}

The main idea of our approach is to bundle sets of synaptic weights and feed them to the DSP48E2 multiplexers and switch the multiplexer with spikes, as illustrated in Fig.\ref{fig:dsp}C.
In this work, the synaptic connection weights are quantized into INT8 by the well-established post-training quantization or quantization-aware training methods developed in traditional neural networks (NNs).
Four sets of INT8 weights are signed extended to INT12 and concatenated into 48-bit. The upper 30 bits are assigned to the input port A while the lower 18 bits are assigned to input port B.
As shown in Fig.\ref{fig:dsp}C, $w_{20},w_{21},w_{22},w_{23}$ are bundled and assigned to port A and B. A and B are then concatenated and multiplexed by the X multiplexer.
In SNNs, the input spikes are shared by different sets of weights through the axons, as shown in Fig.\ref{fig:dsp}A. In this case, spike $s_2$ is fetched to dynamically switch the X multiplexer between the four sets of weights (A:B) and all 0s.
Similarly, another four sets of INT8 weights, $w_{30},w_{31},w_{32},w_{33}$, are signed extended, concatenated, and directly assigned to the C data input.
Another spike, $s_3$, is fetched to dynamically switch the W multiplexer between C and all 0s.
The OPMODE is dynamically controlled by $s_2$ and $s_3$.
The Y multiplexer outputs are set to all 0s.
The Z multiplexer selects the PCIN inputs and the partial sum from the lower DSP slice. The results are staged into the P register and propagated to the upper DSP slice through the PCOUT data port.
Therefore, the arithmetic function of a single DSP48E2 slice is equivalent to a $2 \times 4$ synaptic crossbar computation without general fabric logic overhead.

% Mathematically, the arithmetic function of a single DSP48E2 slice configured to support synaptic operation can be expressed as:

% \begin{equation*}
%     \vect{p_{i}}=\vect{s_i}\cdot \matr{W_i}+\vect{p_{i-1}},\vect{p_{-1}}=\vect{0}.
%     \end{equation*}
    
%where $\bm{s_i}$ is the $1\times 2$ binary spike vector, and $\bm{w_i}$ is the $2\times 4$ INT8 synaptic weights matrix, $\bm{p_i}$ is the $1\times 4$ partial sum vector, and the $\bm{p_{i-1}}$ is the partial sum vector contributed by the lower DSP slice with the same shape as $\bm{p_i}$. $\cdot $ represents the spikes-weights vector-matrix multiplication.

\subsubsection{\bigchange{Synaptic Crossbar Computation by a DSP48E2 Chain}}

There are dedicated internal paths between adjacent DSP48E2s for local cascading, which will not occupy global routing resources. Since the synaptic weights are quantized to 8 bits and the bit width of the SIMD adder is 12 bits, up to 8 DSP48E2 can be cascaded in a chain without numeric overflow.
While the multiplexer and the ALU unit in the DSP48 are used for the synaptic crossbar computation, the dedicated cascaded path of the DSP48E2 acts as the dendrite, collecting and accumulating the computation results along the DSP48E2 chain.
In this way, the arithmetic function of a DSP48 cascaded chain of length 8 is equivalent to a $16 \times 4$ synaptic crossbar computation.
A DSP48 cascaded chain of length 8 is a single PE (Processing element) in FireFly, which is the basic element of the systolic array introduced in the next subsection.

The straightforward implementation of a $2 \times 4$ synaptic crossbar computation described above using general fabric will consume 86 Look-up-tables, 114 Flip-flops and 8 Carry chains, while a $16 \times 4$ crossbar will consume 688 Look-up-tables, 912 Flip-flops and 64 Carry chains, shown in Table.\ref{resutil}.
Note that general fabric implementation will also consume global routing resources. It is considerably less efficient than the proposed approach and will lead to a compromised clock frequency when the parallelism scales up.

% Similarly, the arithmetic function of a DSP48E2 chain of length 8 configured to support synaptic operation can be mathematically expressed as below:

% \begin{equation*}
%     \vect{p}=\sum_{i=0}^{7} \vect{s_i}\cdot \matr{W_i}=\vect{s}\cdot \matr{W}.
% \end{equation*}

%where $\bm{s}$ is the $1\times 16$ binary spike vector, and $\bm{W}$ is the $16\times 4$ 8-bit-integer (INT8) synaptic weights matrix, $\bm{p}$ is the $1\times 4$ partial sum vector.

% \begin{table}[]
%     \begin{center}
%     \caption{Resource Utilization Comparison.}
%     \label{resutil}
%     \begin{tabular}{c|c|c|c|c}
%     \hline
%             & DSP48E2 & LUT & FF  & CARRY8 \\ \hline
%     DSP    & 1       & 0   & 0   & 0      \\ \hline
%     Fabric & 0       & 86  & 114 & 8      \\ \hline
%     \end{tabular}
%     \end{center}
% \end{table}

\begin{table}[]
\begin{center}
\caption{Resource Utilization Comparison Between DSP and Fabric Implementation.}
\label{resutil}
\begin{tabular}{c|c|c|c|c|c}
\hline
\multicolumn{1}{r|}{Crossbar Size} & Tech   & DSP48E2 & LUT & FF  & CARRY8 \\ \hline\hline
\multirow{2}{*}{$2\times 4$}               & DSP    & 1       & 0   & 0   & 0      \\ \cline{2-6} 
                                & Fabric & 0       & 86  & 114 & 8      \\ \hline
\multirow{2}{*}{$16 \times 4$}              & DSP    & 8       & 0   & 0   & 0      \\ \cline{2-6} 
                                & Fabric & 0       & 688 & 912 & 64     \\ \hline
\end{tabular}
\end{center}
\end{table}

\subsubsection{\bigchange{Synaptic Crossbar Computation by a Systolic Array}}

By instantiating multiple DSP48E2 chains, or PEs, in a systolic array fashion (shown in Fig.\ref{fig:dsp}D), we can support a larger synaptic crossbar. The systolic array is a specialized mesh of homogeneous PEs designed to process massive parallel computations. It has the potential to run at a high frequency due to its regular and adjacent interconnections. \minorchange{Previous FPGA neuromorphic hardware adopting a systolic array architecture failed to achieve satisfactory performance, either in resource efficiency or clock frequency, since they are implemented in low-speed general fabrics.} FireFly makes full use of the DSP48E2 feature and greatly improves the systolic array's performance. 

In this work, our definition of the systolic array size is the same as that of the synaptic crossbar. A $M\times N$ systolic array support a $M \times N$ synaptic crossbar computation, consisting of $\frac{M}{16} \times \frac{N}{4}$ PEs, or $\frac{M}{16} \times \frac{N}{4} \times 8$ DSP48E2s. Note that the DSP48E2 chain acting as the dendrite in each PE cannot be cascaded across PEs without numeric overflow, therefore four additional adder-trees are instantiated to sum the SIMD accumulating results from $\frac{M}{16}$ PEs at each column up.

Each PE in the systolic array contains different sets of synaptic weights. Adopting a weight-stationary scheme, the synaptic weight matrix remains cached in a PE until they are no longer needed. The same $1\times M$ binary spike vector is shared across columns horizontally behaving just like the axons. The $1\times N$ partial sums, or the synaptic currents, flow out of the systolic array vertically behaving just like the dendrites.

% Mathematically, the arithmetic function of the systolic array can be expressed by a matrix-vector multiplication as below:

% \begin{equation*}
%     \bm{P}=\bm{S}\cdot \bm{W}.
% \end{equation*}

\subsection{Spike Vector Generation for Convolution by Line Buffer}
Similar to ANN, 2-D convolution is the basic operation in a digitalized SCNN. We incorporate the traditional line buffer design to generate the spike window needed for the spike-map convolution. The line buffer is commonly seen in CNN accelerator design because it can efficiently achieve kernel-level parallelism and ensure good reuse of image data.

When FireFly is configured to SCNN mode, $C_{in}$ channels of binary spike map are bundled together and stream into the line buffer. The $K_h \times K_w$ spikes-bundle window is then flattened to a $K_h\times K_w \times C_{in}$ vector and sent to the systolic array. In most of the established CNN architectures, $3\times 3$ convolution with stride 1 and the same padding is the most common configuration. The SCNN architecture follows this scheme. Ideally, general neuromorphic hardware for SNN should support all types of convolutional layers with different configurations. But the hardware would not work efficiently for all types of convolution configuration and such design would cause hardware overhead, thus might not be feasible. Therefore, we design specialized line buffer logic for $3\times 3$ convolution. Nevertheless, the methods discussed here are compatible with other kernel sizes. Using the Dynamic Function Exchange features in FPGA, hardware supporting different types of convolutional layers can be dynamically deployed in FPGA during runtime.

When FireFly is configured for multi-layer perception (MLP) topology mode, the line buffer datapath for SCNN is left idle and the shift register datapath for MLP is switched on. 
The shift register forms a serial-to-parallel stream width adapter by combining the $C_{in}$ input spikes of $K_h \times K_w$ input transactions into one. The length of the binary spike vector in SCNN and MLP datapaths is the same and compatible with the height of the systolic array. 

\subsection{\bigchange{Synaptic Weight Delivery in a Multi-level Hierarchy}}

\begin{figure*}
    \centering
    \includegraphics[width=1.0\linewidth]{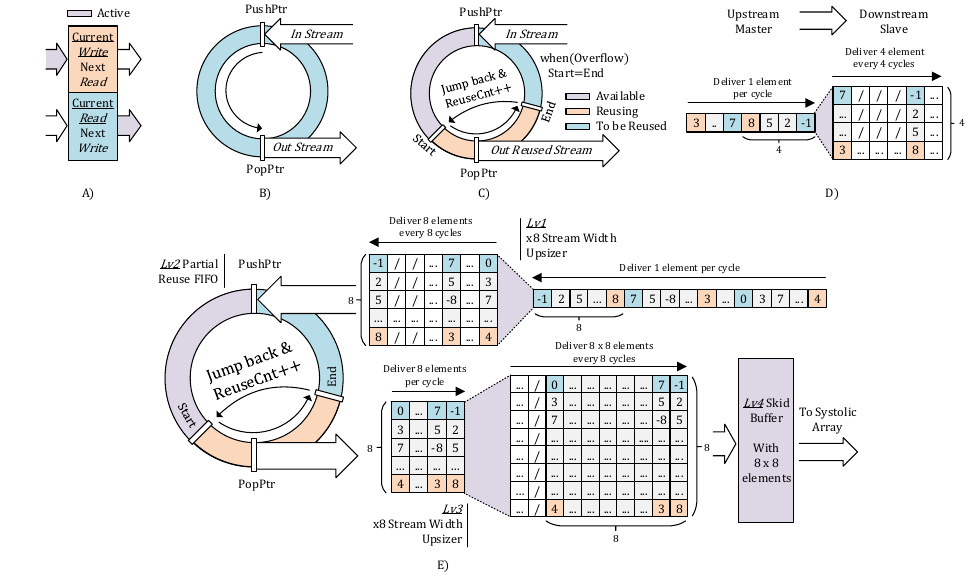}
    \caption{Different Approaches for Hiding Data Transfer Latency to Improve Throughput. A) A Ping-pong buffer. B) A Synchronous FIFO. C) The Proposed Partial Reuse FIFO. \minorchange{D) A $\times 4$ Stream Width Upsizer E) A Four-level synaptic weights delivery hierarchy to enable synaptic weights reusing, reduce off-chip memory bandwidth and hide the weight loading latency to the systolic array.}}
    \label{fig:fifo}
\end{figure*}

Although a DSP48E2-featured systolic array can already support a large synaptic crossbar computation, it is not feasible to build a static synaptic crossbar circuit large enough for SNNs that have millions of neurons and synaptic connections. Instead, the pre-synaptic neurons and post-synaptic neurons should share the same synaptic crossbar computation circuit in a time-multiplexed manner. 

Considering the inference process of a single convolution layer in SCNN, all pre-synaptic neurons within the same channel of the same feature map share the same weight kernel, so the weight matrix can remain static while the input spikes of feature maps flow through the systolic array in a streaming pipeline.
However, the weight matrix needs to be changed after the current subset of feature maps finishes processing. Replacing the current set of weights with the next set of weights can be problematic.
On the one hand, the instantaneous data reloading bandwidth is extremely high when the current set of weights expires.
On the other hand, the expired set of weights at the current timestep will be reloaded again at the next step. It is inefficient if synaptic weights need to be fetched from off-chip memory over and over again at every timestep.

We propose a multi-level weight delivery hierarchy to tackle the aforementioned problems. The instantaneous bandwidth needed when reloading the next set of weights is amortized over an idle period when the weights are kept stationary by the multi-level weight delivery hierarchy. Synaptic weights are cached on-chip and reused over all timesteps using a novel memory structure we proposed to avoid repetitive off-chip data access. As we iterate the tiled output channel variable over timesteps, shown in Algorithm.\ref{algo}, only a small portion of weights contributing to the current subset of output feature maps need to be cached on-chip. Data width upsizing techniques are used to boost the on-chip data bandwidth to enable faster weight delivery.
\minorchange{
There are three basic components in the proposed multi-level weight delivery hierarchy: The proposed Partial Reuse FIFO, the stream width upsizer and the skid buffer.}

\subsubsection{\bigchange{Partial Reuse FIFO}}
We propose Partial Reuse FIFO, a new memory structure for streaming data buffering, supporting data reuse like the Ping-pong buffer and having a FIFO-like feature. 
We first review two classic memory structures for streaming data buffering and latency hiding before we introduce the Partial Reuse FIFO.
\minorchange{Fig.\ref{fig:fifo}A} shows a classic Ping-pong buffer. The buffer size is doubled for independent read and write processes. The input stream flows into one bank of the buffer and the output stream flows out from the other. Read and write conflicts are eliminated but memory resource consumption is relatively high for double buffering. Data cached in the Ping-pong buffer can be reused but manual controlling and bank switching are needed, which may complicate the controller design.
\minorchange{Fig.\ref{fig:fifo}B} shows a classic synchronous FIFO, which is represented using a ring. A push pointer is used to mark the write address of the incoming data stream. A pop pointer is used to mark the read address of the output data stream. When the push pointer and the pop pointer meet each other, the FIFO is either full or empty, depending on whether the occupancy of the FIFO is rising or falling. FIFO provides a certain capability of buffering the input data stream when the downstream module is not ready. The control logic of the FIFO is self-contained. Using a valid-ready handshaking protocol, FIFO can be inserted directly between modules without complicating the whole design. However, FIFO does not support data reusing.

The Partial Reuse FIFO we proposed is shown in \minorchange{Fig.\ref{fig:fifo}C.} The mechanism of the Partial Reuse FIFO is the same as the traditional synchronous FIFO, except that a partial region in the FIFO ring cannot be flushed by incoming data until it is reused $T$ times, where $T$ is a control register of the Partial Reuse FIFO. The reuse region of the FIFO is labeled by $Start$ and $End$. The pop pointer jumps back to the $Start$ position whenever it reaches the $End$. The reuse counter increases whenever the pop pointer jumps back to $Start$. The $Start$ label stays the same when the region is still being reused. When the counter reaches $T$, the counter is reset, label $End$ becomes the next label $Start$ and the next label $End$ is set by $Start+L-1$, where $L$ is another control register of the Partial Reuse FIFO. When the push pointer meets the label $Start$, the Partial Reuse FIFO is considered full and the ready signal to the inputs stream is cleared. When label $End$ is ahead of the push pointer, the Partial Reuse FIFO is considered empty until the reuse sector of the FIFO is filled by the input stream.  Using the valid-ready handshaking protocol, the function of the Partial Reuse FIFO is self-contained, with only two control registers, the reusing times $T$ and the reusing length $L$, exposed. The Partial Reuse FIFO contains only a monolithic RAM and does not need to be double-buffered. The push-pop pointer in the FIFO control logic ensures no read-write collision. The reuse sector protected by the $Start$ and $End$ labels enables data reuse. New data from multiple batches can be pushed to the Partial Reuse FIFO sequentially as long as the FIFO is not full. The Partial Reuse FIFO is the key component in this multi-level synaptic weight delivery hierarchy.

\subsubsection{\bigchange{Stream Width Upsizer}}
A $\times N$ stream width upsizer converts the 1-element input stream to a $N$-element output stream by allocating $N$ elements of the input stream and firing them all at once. As shown in Fig.\ref{fig:fifo}D, the data stream from the upstream master delivers $-1,2,5,8,7...3$ serially, delivering 1 element per clock cycle. The $\times 4$ stream width upsizer performs a serial-to-parallel conversion, delivering 4 elements every 4 clock cycles. The average data throughput of the upstream and downstream measured over time is the same, but the instantaneous throughput is increased $\times 4$ times.
A Partial Reuse FIFO module can be directly placed after the stream width upsizer to boost the data throughput once the reuse sector region of the Partial Reuse FIFO is filled.

\subsubsection{\bigchange{Skid Buffer}}
A skid buffer is a two-entry Pipeline FIFO Buffer. It decouples two sides of a ready/valid handshake to allow back-to-back transfers. The skid buffer is placed at the last stage of the multi-level weight delivery hierarchy. At the downstream side of the skid buffer, the systolic array holds the current set of weights stationary by applying back pressure to the skid buffer and releasing the pressure when the current set of weights is no longer needed. At the upstream side of the skid buffer, the ready signal is always held high until new data shifts in, blocking the back pressure of the systolic array to enable faster data delivery.

\minorchange{Fig.\ref{fig:fifo}E} shows a simple example illustrating the mechanism of the multi-level weight delivery hierarchy.
Arrows indicate the direction of data transfer.
We assume the weight data stream coming from off-chip memory delivers 1 element per clock cycle (for the simplicity of drawing). The multi-level weight delivery hierarchy consists of four levels as listed below.

\minorchange{
1) Level 1: The data stream is upsized by the $\times 8$ stream width upsizer, delivering 8 elements every 8 clock cycles.
As shown in Fig.\ref{fig:fifo}E, $-1,2,5...8$, $7,5,-8...3$, $0,3,7...4$ flow into the upsizer one by one serially. $(-1,2,5...8),(7,-5,-8...3),...,(0,3,7...4)$ flow out of the upsizer 8 elements in a group. Elements with slash symbols are invalid at the current clock cycle.
}

\minorchange{
2) Level 2: The Partial Reuse FIFO is placed right after the upsizer.
Once the data is cached in the reuse region, the output stream of the Partial Reuse FIFO can deliver 8 elements per cycle, thus boosting the data throughput.
Weight Data cached in the reuse region is reused $T$ times before being flushed with new data.
Note that invalid elements no longer occupy clock cycles in the output data stream of the Partial Reuse FIFO as shown in Fig.\ref{fig:fifo}E.
}

\minorchange{
3) Level 3: Another $\times 8$ stream width upsizer is placed following the Partial Reuse FIFO to further expand the instantaneous bandwidth. In this case, $8\times 8$ elements are collected and delivered all at once.
}

\minorchange{
4) Level 4: Finally, the skid buffer is instantiated to bridge the weight delivery logic and the systolic array.
}

The multi-level weight delivery hierarchy enables instant weight data supply to the systolic when the current set of weights expires, minimizing the idle state of the systolic array, thereby greatly raising the ratio between the actual throughput and the theoretical throughput.

\subsection{Psum-Vmem Unified Buffer and Spike Generation Logic}

A classic systolic array consumes data from the input data domain and the weight data domain and generates data for the output data domain. If one data domain stays stationary, the other two must flow through the computing logic. This metric holds for the three classic input, weight and output stationary dataflows.
Our architecture adopts the weight stationary dataflow. In this case, synaptic weights remain stationary in the systolic array, and the input binary spikes and the output flow in and out of the systolic array. The flowing spike vector is generated by the line buffer mechanism, and the outputs are stored in the proposed Psum-Vmem Unified Buffer.

In our architecture, the synaptic operations in SNN are spatially parallelized. However, it is unlikely to flatten a whole layer spatially onto the area-power-restricted hardware substrates. Therefore, certain tiling strategies need to be implemented. We adopt the channel tiling strategy to accommodate layers with a large number of channels to the same systolic array. Input spike map channels are split into multiple tiles to fit into the height of the systolic array. Output spike map channels are calculated $N$ at a time according to the width of the systolic array.
In every single timestep, the partial sums of the $N$ output spike map channels are stored on-chip and are not fully accumulated until all tiles of the input spike map channels are calculated. In each layer, the membrane voltage of the $N$ output spike map channels are also needed to be stored on-chip until all timesteps are iterated. Instead of instantiating a separate buffer for partial sum and membrane voltage, we propose the Psum-Vmem Unified Buffer to reduce RAM consumption.

\begin{figure}
    \centering
    \includegraphics[width=1.0\linewidth]{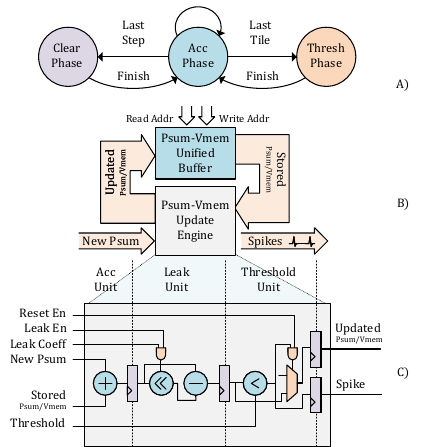}
    \caption{Psum-Vmem Update Mechanism. A) The finite-state machine performing the Psum-Vmem update. B) The proposed Psum-Vmem unified buffer and Psum-Vmem update engine. C) The hardware implementation details of the Psum-Vmem update engine.}
    \label{fig:pbuffer}
\end{figure}

Since tiles of input spike map channels in a single timestep are sent to the computing array one by one and the temporal dimension of SNN is kept in its natural way of executing sequentially, the partial sum accumulating process and the membrane voltage update process can be scheduled using a finite state machine. There are three states specified in the FSM: accumulating phase, thresholding Phase, and clearing phase.
During the accumulating phase, Psum extracted from the Psum-Vmem unified buffer is accumulated by the computing results from the systolic array.
When the last tile of the input spike map channel in the current timestep arrives and the current timestep is not the last, the FSM switches to the thresholding phase. The extracted Psum is first accumulated, then processed by the optional leaky unit and the thresholding unit, and eventually written back to the unified buffer. The accumulated Vmem will be subtracted from a fixed portion of its value by the optional leaky unit to support the LIF neuron dynamics. The thresholding unit will compare the Vmem with the threshold, generate a spike, and reset the Vmem if it exceeds the threshold. All of the computations are pipelined to improve timing. The FSM switches back to the accumulating phase when this phase finishes.
When the last tile of the input spike map channel in the last timestep arrives, the FSM switches to the Clearing Phase. The computation process is the same as the thresholding phase, except that the Vmem value will be cleared to reset the unified buffer for the next SNN layer.

\section{Implementation and Experiments}

\subsection{Experiments Setup}

FireFly is mapped onto several off-the-shelf commercially available Xilinx Zynq Ultrascale FPGAs, including the Ultra96v2, KV260 and ZCU104 evaluation boards. The FPGA chips of the three evaluation boards are xczu3eg, xczu5ev, and xczu7ev, respectively. Most neuromorphic hardware uses expensive large FPGA devices, ignoring the feasibility of deploying such hardware in the real world. FireFly brings hope to SNN real-world applications in an edge scenario.

Our proposed FireFly is designed using SpinalHDL. The Verilog codes generated by the SpinalHDL compiler are synthesized and implemented in the Xilinx Vivado 2021.1 with ML-Based design optimization to achieve a higher clock rate and faster timing closure.
Power consumption estimates and timing results are obtained after place-and-route using the power analysis and timing summary tools in the Vivado Design Suite which provides detailed analysis and accurate estimation.
We choose the Zynq devices as the system platforms. The built-in host CPU controller enables fast deployment of different SNN networks without the need to change the PL logic. The host program generates a command sequence in advance and sends the commands to PL through a high-performance AXI-Stream to the internal command queue of the AXI DataMover.
%Throughput performance is obtained by recording the timer value on the PS side of Zynq while the PL runs the benchmark tasks.
\minorchange{
FireFly is based on the Brain-inspired Cognitive Engine (BrainCog) and is a first step towards the software-hardware co-design for the BrainCog project (\url{http://www.brain-cog.network/})\cite{zeng2022braincog}.
}

\begin{table*}[]
    \centering
    \begin{threeparttable}[b]
    \caption{Comparison with other works in Hardware Specifications.}
    \begin{tabular}{c|c|c|c|c||c|c|c}
    \hline
    & \begin{tabular}[c]{@{}c@{}}Cerebron\\ TVLSI'22\cite{chen2022cerebron}\end{tabular} & \begin{tabular}[c]{@{}c@{}}SIES\\ JCST'20\cite{wang2020sies}\end{tabular} & \begin{tabular}[c]{@{}c@{}}Ye et al.\\ TCAD'22\cite{ye2022implementation}\end{tabular} & \begin{tabular}[c]{@{}c@{}}Guo et al.\\ GLSVLSI'19\cite{guo2019systolic}\end{tabular} & \textbf{Ours}\tnote{6}           & \textbf{Ours}\tnote{7}           & \textbf{Ours}\tnote{8}                     \\ \hline\hline
    Device             & xc7z100                                                     & xcvu440                                                & xc7k325t                                                    & xc7vx690t                                                       & \textbf{xczu3eg}        & \textbf{xczu7ev}        & \textbf{xczu5ev}                  \\ \hline
    SoC                & 7 series                                                    & No                                                     & No                                                          & No                                                              & \textbf{Ultrascale}     & \textbf{Ultrascale}     & \textbf{Ultrascale}               \\ \hline
    Dataflow           & OS                                                          & OS                                                     & OS                                                          & OS                                                              & \textbf{WS}             & \textbf{WS}             & \textbf{WS}                       \\ \hline
    Precision          & /                                                           & FIX32                                                  & FIX16                                                       & FIX32                                                           & \textbf{INT8}           & \textbf{INT8}           & \textbf{INT8}                     \\ \hline
    Neuron             & IF                                                          & IF                                                     & EPC-LIF                                                     & IF                                                              & \textbf{IF/LIF}         & \textbf{IF/LIF}         & \textbf{IF/LIF}                   \\ \hline
    Array Size         & $8\times 8\times 4$\tnote{1}                                         & $64\times 64$                                          & $16\times 16$                                               & $32\times 32$                                                   & \textbf{$144\times 16$} & \textbf{$288\times 32$} & \textbf{$144\times 16\times   2$} \\ \hline
    Mux-Acc Unit       & 256                                                         & 4096                                                   & 256                                                         & 1024                                                            & \textbf{2304}           & \textbf{9216}           & \textbf{2304$\times$2}            \\ \hline
    Frequency(MHz)     & 200                                                         & 200                                                    & 200                                                         & 100                                                             & \textbf{300}            & \textbf{300}            & \textbf{300}                      \\ \hline
    Peak GOP/s         & 650.0\tnote{2}                                                       & 1638.4\tnote{3}                                                 & 102.4\tnote{4}                                                       & 204.8\tnote{5}                                                           & \textbf{1382.4}         & \textbf{5529.6}         & \textbf{1382.4$\times$2}          \\ \hline\hline
    Available LUT(K)   & 277                                                         & 2532                                                   & 203                                                         & 433                                                             & \textbf{70}             & \textbf{230}            & \textbf{117}                      \\ \hline
    Used LUT(K)        & 86                                                          & 302                                                    & 16                                                          & 53                                                              & \textbf{15}             & \textbf{42}             & \textbf{32}                       \\ \hline
    LUT Utilization    & 31.05\%                                                     & 11.93\%                                                & 7.88\%                                                      & 12.24\%                                                         & \textbf{21.43\%}        & \textbf{18.26\%}        & \textbf{27.35\%}                  \\ \hline
    Available DSP      & 2020                                                        & 2880                                                   & 840                                                         & 3600                                                            & \textbf{360}            & \textbf{1768}           & \textbf{1248}                     \\ \hline
    Used DSP           & 0                                                           & 0                                                      & 0                                                           & 0                                                               & \textbf{288}            & \textbf{1152}           & \textbf{576}                      \\ \hline
    LUT Utilization    & 0.00\%                                                      & 0.00\%                                                 & 0.00\%                                                      & 0.00\%                                                          & \textbf{80.00\%}        & \textbf{65.16\%}        & \textbf{46.15\%}                  \\ \hline
    Available B/URAM   & 755                                                         & 2520                                                   & 445                                                         & 1470                                                            & \textbf{216}            & \textbf{312/96}         & \textbf{144/64}                   \\ \hline
    Used B/URAM        & 283                                                         & 192                                                    & 220                                                         & 65                                                              & \textbf{162}            & \textbf{25/40}          & \textbf{16/24}                    \\ \hline
    B/URAM Utilization & 37.48\%                                                     & 7.62\%                                                 & 49.44\%                                                     & 4.42\%                                                          & \textbf{75.00\%}        & \textbf{11.5/41.6\%}    & \textbf{11.1/37.5\%}              \\ \hline
            \end{tabular}
    \label{tab:gops}
    \begin{tablenotes}
    \item[1] Cerebron comprises $8 \times 8$ computing units (CUs), each of which houses 4 PEs.
    \item[2] The theoretical GOP/s of Cerebron is 102.4 as calculated by Equ.\ref{peakgop}. However, by leveraging a two-step weight sparsity acceleration technique, Cerebron achieved an improved GOP/s of 650.0, which is presented in the table for a fair comparison.
    \item[3] The theoretical GOP/s of SIES is 1638.46 according to Equ.\ref{peakgop}. Nonetheless, SIES reported a GOP/s of 1562.5, even though the definition of peak GOP/s in \cite{wang2020sies} is identical to Equ.\ref{peakgop}. The higher GOP/s is provided in the table for a fair comparison.
    \item[4] Although not reported in \cite{ye2022implementation}, the GOP/s is calculated using Equ.\ref{peakgop}.
    \item[5] Although not reported in \cite{guo2019systolic}, the GOP/s is calculated using Equ.\ref{peakgop}.
    \item[6] FireFly employs a $144 \times 16$ systolic array and is implemented on Ultra96v2.
    \item[7] FireFly employs a $288 \times 32$ systolic array and is implemented on ZCU104.
    \item[8] FireFly comprises two $144 \times 16$ systolic arrays and is implemented on KV260.

\end{tablenotes}
    \end{threeparttable}
\end{table*}

\subsection{\bigchange{Comparisons in Hardware Specifications}}

The theoretical peak GOP/s of an SNN accelerator is given as:
\begin{equation}
    \label{peakgop}
    \text{Peak GOP/s}=2\times f\times S.
\end{equation}

where $f$ is the system clock frequency, and $S=M \times N$ denotes the size of the systolic array. The peak GOP/s calculation is the same as \cite{wang2020sies} and \cite{chen2022cerebron}. In FireFly, $M$ denotes the number of rows in the systolic array, while $N$ denotes the columns. The peak performance should be proportional to the systolic array size.
The size of the systolic can be statically reconfigured in FireFly according to the on-chip resources on different evaluation boards. A $M\times N$ systolic array in FireFly receives $M$ presynaptic inputs and produces partial sum for $N$ neurons, where $N=P$ and $M=K_h\times K_w \times P$. The resource consumption, memory bandwidth and acceleration performance are linearly proportional to the parallelism factor $P$. $P$ can be any value as long as the systolic array can fit in the target device. As $P$ is also the tiling factor of the input and output channels in a convolutional layer, it is preferable to set $P$ to a power of 2 because the number of channels in most convolutional layers is a power of two. Therefore, we evaluate two representative configurations, $144\times 16$ and $288\times 32$ to demonstrate the reconfigurability of FireFly.
Implementing synaptic operations using DSP48 significantly reduces fabric overhead and leads to substantial improvements in GOP/s compared to most existing hardware. FireFly, with a $144\times 16$ systolic array, can achieve a peak performance of 1382.4 GOP/s, while FireFly with a $288\times 32$ systolic array can achieve a peak performance of 5529.6 GOP/s, as presented in Table \ref{tab:gops}.

\minorchange{
We compare with four representative systolic-array-based hardware accelerators implemented on FPGA platforms in Table \ref{tab:gops}.
We focus on comparing the hardware specifications and theoretical computing capabilities of these accelerators, which can be easily quantified.
As these accelerators are based on a systolic array, the regular 2D arrangement of PEs in these works makes it simple to measure the maximum computing power that these accelerators can deliver.
Despite variations in their PE configurations, all designs employ a basic multiplex-accumulate unit to implement the synaptic crossbar computation.
The peak throughput is determined by the number of multiplex-accumulate units and the clock rate and can be estimated using Equation \ref{peakgop}.
}

\minorchange{
We first present a comprehensive perspective by outlining several essential observations.
1) Despite the abundance of DSP resources in their devices, none of these works effectively utilize them, resulting in high LUT consumption and low clock frequency.
2) It's worth noting that the FPGA devices used in these works, such as xc7z100, xcvu440, xc7k325t, and xc7vx690t, are considerably larger than the edge device we employed (xczu3eg), yet we were able to build a larger computing array and achieve comparable, or even better, peak performance.
3) Although these works employ expensive, large FPGA devices, they do not effectively harness the full potential of these resources and fail to consider the practical feasibility of real-world deployment.
}

\minorchange{
We then make case-by-case comparisons with these four representative works, since finding a normalized metric that considers all aspects, such as precision, neuron types, and resource consumption, can be extremely challenging.
}

\minorchange{
1) Cerebron\cite{chen2022cerebron} leverages weight sparsity acceleration and supports pointwise and depthwise convolutions, resulting in a more complex processing element (PE) design than that of FireFly. However, FireFly achieves a higher peak performance (1382.4 vs 650 GOP/s) using a smaller device (xczu3eg vs xc7z100), surpassing Cerebron in terms of computational density efficiency. A drawback of FireFly is its inability to benefit from weight sparsity acceleration or support pointwise and depthwise convolutions.
}

\minorchange{
2) SIES\cite{wang2020sies} utilizes a $64\times 64$ systolic array, which provides a peak performance of 1638.4 GOP/s on the xcvu440 platform. Guo et al. \cite{guo2019systolic} implement a $32\times 32$ systolic array, delivering a peak performance of 204.8 GOP/s. It is worth noting that both SIES and Guo use FIX32 precision, while FireFly uses INT8. To ensure a fair comparison, we use Tb/s (Tera bit-operations per second) to account for data precision.
We compare FireFly, mapped on xczu3eg, to Guo's implementation and the results show that FireFly surpasses Guo's implementation. (11.1 vs 6.5)
We compare FireFly, mapped on xczu7ev, to SIES and the results show that FireFly slightly trails behind SIES (44.24 vs. 52.4). 
However, FireFly is more efficient in terms of resource utilization, with a balanced LUTs and DSPs consumption compared with both implementations.
}

\minorchange{
3) Ye et al. \cite{ye2022implementation} implement a systolic array that supports accurate LIF dynamics using an extended prediction correction technique, while FireFly only approximates LIF behavior using a simple shift operation. Ye et al. support MLP and CNN topologies using separate computing units (PE arrays + FC cores), while FireFly manages to reuse the same systolic array. Although Ye et al. cannot achieve comparable computing throughput with FireFly, they can deliver a more precise LIF behavior.
}

% To the best of our knowledge, SIES\cite{wang_sies_2020} achieves the highest GOP/s among all the existing FPGA-based accelerators. Compared with SIES\cite{wang_sies_2020}, FireFly mapped on xczu3eg consumes only $\frac{1}{20}$ LUTs and $\frac{1}{8}$ FFs but still achieve similar GOP/s, whereas FireFly mapped on xczu7ev consumes only $\frac{1}{7}$ LUTs and $\frac{1}{2}$ LUTs FFs and achieves a $\times 3.5$ speed up. Additionally, we map two heterogeneous FireFly cores onto xczu5ev to support the concurrent inference of two independent SNNs.

% We can still achieve higher throughput when compared with SpinalFlow and SATO, which are state-of-the-art SNN hardware accelerators built in 28nm ASIC. We are well aware that it is difficult to make an apples-to-apples comparison with the hardware adopting different design methodologies, supporting different types of neurons, using different synaptic weight precisions or implementing on different platforms, FireFly can still be called a high-performance SNN accelerator due to its excellent GOP/s performance.

\subsection{\bigchange{Comparisons in Benchmark Evaluations}}

\begin{table*}[]
    \centering
    \begin{threeparttable}[b]
    \caption{Comparison with Related Work for Multiple Image Classification Tasks Using SNNs for Multiple Datasets.}
    \begin{tabular}{c|c|c|c|c|c|c|c|c|c}
    \hline
                                   & Network                                                                  & Datasets              & Acc.           & MFLOPS          & FPS           & Power         & kFPS·MFLOPS     & Eff.            & Device                            \\ \hline\hline
    Minitaur\cite{neil2014minitaur}                          & 784-500-500-10                                                           & MNIST                & 94.2           & 1.29            & 108           & 1.5           & 0.14            & 0.09            & xc6slx150t                        \\ \hline
    Han et al.\cite{han2020hardware}                         & 784-1024-1024-10                                                         & MNIST                & 97.06          & 3.72            & 161           & 0.477         & 0.60            & 1.26            & xc7z045                           \\ \hline
    Zhang et al.\cite{zhang2019asynchronous}                 & 784-512-384-10                                                           & MNIST                & 98             & 1.20            & 909           & 0.36          & 1.09            & 3.04            & xc7vx690t                         \\ \hline
    Ju et al.\cite{ju2020fpga}                                 & \begin{tabular}[c]{@{}c@{}}28x28-64c5-p2-64c5-p2-\\ 128-10\end{tabular}  & MNIST                & 98.94          & 15.21           & 164           & 4.6           & 2.50            & 0.54            & xczu9eg                           \\ \hline
    Fang et al.\cite{fang2020encoding}                       & \begin{tabular}[c]{@{}c@{}}28x28-32c3-p2-32c3-p2-\\ 256-10\end{tabular}  & MNIST                & 99.2           & 4.87            & 133           & 4.5           & 0.65            & 0.14            & xczu9eg                           \\ \hline
    \multirow{2}{*}{Ye et al.\cite{ye2022implementation}}    & 784-512-256-128-64-10                                                    & FMNIST               & 89.01          & 1.15            & 7142          & 0.699         & 8.20            & 11.73           & \multirow{2}{*}{xc7k325t}         \\ \cline{2-9}
                                   & \begin{tabular}[c]{@{}c@{}}32x32-32c3-p2-32c3-p2-\\ 256-10\end{tabular}  & SVHN                 & 82.15          & 6.36            & 826.4         & 0.982         & 5.26            & 5.35            &                                   \\ \hline
    \multirow{2}{*}{Cerebron\cite{chen2022cerebron}}         & 28x28-16c3-32c3-10                                                       & MNIST                & 99.4           & 7.95            & 38500         & 1.4           & 306.15          & 218.68          & \multirow{2}{*}{xc7z100}          \\ \cline{2-9}
                                   & MobileNet\tnote{1}                                                                & CIFAR10              & 91.9           & 1179.65         & 90            & 1.4           & 106.17          & 75.83           &                                   \\ \hline
    \multirow{3}{*}{E3NE\cite{gerlinghoff2021e3ne}}          & LeNet5                                                                   & MNIST                & 99.1           & 11.71           & 3400          & 3.4           & 39.81           & 11.71           & \multirow{3}{*}{xcvu13p}          \\ \cline{2-9}
                                   & AlexNet\tnote{2}                                                                  & CIFAR10              & 80.6           & 284.16          & 14.3          & 4.7           & 4.06            & 0.86            &                                   \\ \cline{2-9}
                                   & VGG11                                                                    & CIFAR100             & 65             & 586.14          & 6.1           & 5             & 3.58            & 0.72            &                                   \\ \hline
    \multirow{3}{*}{SyncNN\cite{panchapakesan2022syncnn}}    & \begin{tabular}[c]{@{}c@{}}28x28-32c3-p2-32c3-p2-\\ 256-10\end{tabular}  & MNIST                & 99.3           & 4.87            & 13068         & 0.4\tnote{3}           & 63.62           & 159.04          & \multirow{3}{*}{xczu9eg}          \\ \cline{2-9}
                                   & \begin{tabular}[c]{@{}c@{}}28x28-32c5-p2-64c5-p2-\\ 2048-10\end{tabular} & MNIST                & 99.6           & 11.71           & 1629          & 0.4           & 19.08           & 47.69           &                                   \\ \cline{2-9}
                                   & VGG13                                                                    & CIFAR10              & 90.79          & 456.91          & 62            & 0.4           & 28.33           & 70.82           &                                   \\ \hline\hline
    \multirow{5}{*}{\textbf{Ours}} & \textbf{SCNN-5}\tnote{4}                                                          & \textbf{MNIST}       & \textbf{98.12} & \textbf{130.53} & \textbf{2036} & \textbf{2.55} & \textbf{265.76} & \textbf{104.22} & \multirow{5}{*}{\textbf{xczu3eg}} \\ \cline{2-9}
                                   & \textbf{SCNN-7}\tnote{5}                                                          & \textbf{CIFAR10}     & \textbf{91.36} & \textbf{284.16} & \textbf{966}  & \textbf{2.55} & \textbf{274.49} & \textbf{107.64} &                                   \\ \cline{2-9}
                                   & \textbf{SCNN-11}\tnote{6}                                                         & \textbf{CIFAR100}    & \textbf{64.28} & \textbf{586.14} & \textbf{470}  & \textbf{2.55} & \textbf{275.49} & \textbf{108.03} &                                   \\ \cline{2-9}
                                   & \textbf{SCNN-9}\tnote{7}                                                          & \textbf{DVS-CIFAR10} & \textbf{72.4}  & \textbf{978.43} & \textbf{282}  & \textbf{2.55} & \textbf{275.92} & \textbf{108.20} &                                   \\ \cline{2-9}
                                   & \textbf{SCNN-9}                                                          & \textbf{DVS-Gesture} & \textbf{89.29} & \textbf{978.43} & \textbf{282}  & \textbf{2.55} & \textbf{275.92} & \textbf{108.20} &                                   \\ \hline
    \end{tabular}
    \label{tab:cmp}

    \begin{tablenotes}
    \item[1] Cerebron uses MobileNet to demonstrate its hardware performance, which decomposes the original 3x3 convolution layer into pointwise and depthwise convolutional layers, resulting in a significant reduction in model complexity. However, FireFly does not support pointwise and depthwise convolution. To provide a fair comparison, the FLOPS number of the MobileNet listed in the table is shown as an equivalent model with a 3x3 kernel, which results in a higher FLOPS number than the original MobileNet.
    \item[2] While E3NE claims to employ AlexNet for the CIFAR10 dataset, the original AlexNet supports a 224x224 image resolution, while CIFAR10 uses a resolution of 32x32. The revised AlexNet structure is not shown in E3NE. To ensure a fair comparison, we use the FLOPS number of SCNN-7, which is expected to be no less than the FLOPS of the revised AlexNet for CIFAR10.
    \item[3] SyncNN calculates power usage by recording the power (24.5 W) when the FPGA board is running and subtracting the static power recorded when the FPGA board is idle (24.1 W). In contrast, we believe that most existing research derives power metrics directly from the Vivado report.
    \item[4] SCNN-5: 28x28-16c3-64c3-p2-128c3-p2-256c3-256c3-10
    \item[5] SCNN-7: 32x32-16c3-64c3-p2-128c3-128c3-p2-256c3-256c3-p2-512c3-10
    \item[6] SCNN-11: 32x32-16c3-64c3-64c3-p2-128c3-128c3-128c3-p2-256c3-256c3-256c3-p2-512c3-512c3-100
    \item[7] SCNN-9: 48x48-16c3-64c3-64c3-p2-128c3-128c3-p2-256c3-256c3-p2-512c3-512c3-10
    \end{tablenotes}

    \end{threeparttable}
    \end{table*}

Existing SNN training methods can be categorized into three types: 
Biologically plausible methods are mainly inspired by the synaptic learning rules in the human brain. Spike-timing-dependent plasticity and Hebbian learning rules are extensively used in these methods. Although these methods are energy efficient and biologically plausible, they only work well in shallow networks and toy datasets like MNIST.
Conversion methods convert the analog values of ANNs into the firing rates of SNNs. Although higher accuracy can be achieved by this method, the timestep is too long thus leading to high energy consumption.
Backpropagation algorithms are also introduced into the SNN domain. 
Surrogate gradient helps SNNs perform backpropagation through time (BPTT) so that SNNs can be adopted to larger-scale network structures on more complex datasets.

In this work, we deploy several state-of-the-art SNN networks trained by backpropagation algorithms\cite{shen2022backpropagation} on FireFly to test the inference performance.
\minorchange{
We evaluate not only the static datasets such as MNIST, CIFAR10 and CIFAR100 but also the neuromorphic datasets such as DVS-CIFAR10 and DVS-Gesture.
The models are trained using the Pytorch framework with NVIDIA A100 graphic processing unit(GPU). AdamW algorithm\cite{loshchilov2017decoupled} is used as the optimizer. The learning rate is set to $1\times 10^{-3}$. The membrane potential threshold is set to $0.5$. The membrane time constant $\tau _m$ is set to $2.0$ for LIF neuron models. The batch size is set to 128. The training epochs are set to 600. The models are trained using surrogate functions like quadratic gate and arctangent gradient to overcome the problem of the non-differentiability of binary spikes.}
Direct coding is used to reduce the total timesteps of the SNNs. In our experiment, the timesteps are scaled down to four without a significant accuracy drop. \minorchange{These training algorithms are provided in BrainCog's infrastructures\cite{zeng2022braincog}\cite{braincogweb}.}

After the training process, several steps are needed to deploy the Pytorch-Trained SNN model to FireFly.
We first apply batch norm fusion to merge the batch normalization layer with the preceding convolutional layer to reduce computation complexity.
Then we observe the distribution of the synaptic weights of each layer, calculate the scaling factor and convert the synaptic weights represented in 32-bit floating point numbers to 8-bit signed integers.
The threshold is also quantized using the scaling factor derived from the weight observations.
Note that the performance drop of post-training quantization without further retraining or fine-tuning is negligible in SNN because no scaling errors of multiplications are introduced.

FireFly shows reconfigurability on different SNN models for different image classification tasks. We evaluate four different SNN model structures with 5, 7, 9, and 11 convolutional layers on five different datasets, shown in Table \ref{tab:cmp}.
\minorchange{As different research studies use varying benchmarks that differ in scale and complexity, deriving meaningful comparison results can be challenging.
To ensure a fair comparison among experiments, we have calculated the FLOPS number of the equivalent ANN model of each SNN model to quantify the benchmark size. Please note that we have ignored the timestep of the SNN in our FLOPS calculation. This is because these accelerators may employ a spike aggregation techniques\cite{panchapakesan2022syncnn} to reduce the complexity of their SNN models, or only process timesteps with spikes\cite{neil2014minitaur}, or adopt temporal coding to ensure that only one spike occurs in all timesteps\cite{narayanan2020spinalflow,liu2022sato}, or simply not report the total timesteps used. Therefore, it is not possible to have a fair comparison considering all these aspects. FLOPS of equivalent ANN models can at least provide a rough estimate of the benchmark size. Specifically, we calculate the FLOPS of a single convolutional or fully-connected layer as follows:
}

\begin{gather}
    \label{convflops}
    \text{FLOPS}_{Conv} = 2\times K_h \times K_w \times H \times W \times C_{out} \times C_{in}\\
    \label{mlpflops}
    \text{FLOPS}_{MLP} = 2\times C_{out} \times C_{in}
\end{gather}

\minorchange{
To evaluate the inference performance of these accelerators, we use the metric kFPS·MFLOPS, where kFPS (kilo frames per second) is reported in each experiment and MFLOPS is calculated. To evaluate the efficiency of these accelerators, we divide the kFPS·MFLOPS by power, where power is also reported in each experiment. These two metrics effectively reflect the accelerators' efficiency while taking the benchmark size and power consumption into account.
}

\minorchange{
After analyzing Table.\ref{tab:cmp}, we have made several key observations:
1) FireFly can adapt to various datasets and models and achieve comparable accuracy with other works across all five datasets.
2) FireFly can support deep and large SNN networks, as indicated by the larger MFLOPS values shown in the table. In terms of inference latency without considering the benchmark size, FireFly achieves a moderate level of performance. In terms of power consumption alone, FireFly is not particularly outstanding.
3) FireFly can achieve high kFPS·MFLOPS when taking the benchmark size into account. Only Cerebron's experiment on MNIST using a small ConvNet can surpass FireFly in this regard.
4) FireFly can achieve high computational efficiency compared to most research, especially when the benchmark size is increased. While Cerebron and SyncNN can achieve high efficiency when the benchmark size is small, their performance degrades rapidly when switching to large-size networks.
5) FireFly exhibits a stable kFPS·MFLOPS across various benchmark sizes and datasets, demonstrating its scalability and reconfigurability.
}

Note that our chosen device, xczu3eg, is an edge device having the fewest resources among all the listed hardware, but still, FireFly shows significant improvement in all these benchmarks. When using a larger xczu7ev device, all the inference performances listed above are improved by $\times 4$ because xczu7ev supports higher parallelism and has a peak performance of 5.523 TOP/s. Our system also supports multiple heterogeneous cores running different SNN models concurrently. When targeting xczu5ev, two FireFly cores can be deployed independently to support multiple real-world tasks.

\subsection{Discussion}

We argue that for FPGA-based SNN accelerator design, the benefits of designing complicated hardware supporting spike sparsity may not make up for the losses of irregular interconnect and underutilization of the dedicated hard block. 

The system clock frequency can have a significant impact on inference performance. Compared with ASICs, routing in FPGAs contributes more delay time since logic elements are connected through a series of switching matrices instead of direct physical wires. A complex digital design with irregular interconnect can easily violate the timing requirements even in the most state-of-the-art FPGA devices. Most existing FPGA-based SNN accelerators can only satisfy the timing requirement of at most 200MHz even on the expensive Virtex Ultrascale+ device.
Another important aspect of FPGA low-power system design is to utilize the existing dedicated hard block rather than build one from scratch. Implementing the same function using the dedicated hard block in FPGAs usually consumes less energy than using the general fabric counterparts. However, most existing FPGA-based SNN accelerators fail to delve into the features provided by the existing dedicated hard block and adopt a no-brainer implementation of spike computation using low-speed fabric.

In this paper, FireFly provides a different perspective on designing dedicated neuromorphic hardware for spiking neural networks targeting FPGA devices. We are well aware that it is important to design hardware that supports sparsity acceleration. However, to our best knowledge, few studies \cite{narayanan2020spinalflow}\cite{liu2022sato} targeting ASICs can show significant speed-ups considering this inherent nature of SNNs, not to mention the large majority of FPGA-based designs. Instead of designing complicated circuits to support the sparsity acceleration, FireFly consists of a monolithic systolic array.
% and adopts a straightforward weight stationary dataflow.
The acceleration comes from the clock frequency improvement brought by the regular and simple interconnect of the systolic array, the pipelined arithmetic computations, and, most importantly, the flexible use of the multi-function DSP48E2s.

In fact, the potential of the DSP48E2 is still far from being fully realized. Wu et al. \cite{wu2017high} proposed a high-throughput processing array for matrix multiplication based on DSP supertile and achieved peak DSP clock rates on Xilinx UltraScale (741 MHz) and UltraScale+ (891 MHz) devices. SNN accelerators can incorporate the DSP supertile design and achieve even higher performance.
The potential of other dedicated hard blocks on FPGA is also yet to be exploited. Scaling the Cascades\cite{samajdar2019scaling} fully utilized the dedicated cascade interconnect of the DSP48E2, BRAM36K, and URAM288K and achieved nearly 100 $\%$ usage of these hard blocks, delivering incredible inference speed on MLPerf benchmarks. It is necessary to migrate the existing hardware optimization techniques of ANN accelerator design to SNN neuromorphic hardware research.
Nevertheless, we agree that ideally, the main advantage of new SNN accelerators compared to ANNs on digital hardware comes primarily from exploiting the sparsity of spikes and not from the replacement of MAC operations with AC operations\cite{dampfhoffer2022snns}. Future neuromorphic hardware design should exploit spike sparsity and migrate existing FPGA optimization techniques simultaneously.

\minorchange{
Also, we admit that since we come up with a DSP optimization technique that is tightly coupled to the FPGA device family we are using, it will limit the portability of our Verilog codes and make it difficult for us to convert our design into an ASIC. However, we argue that FPGA-specific optimizations are still necessary for SNN accelerator design.
%As an emerging research field, SNNs' connection topologies, neuron types, encoding schemes and training methods are ever-changing.
As an emerging research field, SNNs' variants are ever-changing.
FPGA implementations of fast-evolving SNN algorithms are preferred over ASIC implementations because of their reconfigurability and flexibility. High-quality FPGA accelerators with FPGA-specific optimizations can offer feasible solutions to SNN real-world applications.
}

\section{Conclusions}
In this work, we introduced a high-throughput and reconfigurable hardware accelerator for spiking neural networks. To achieve high-performance inference of SNN, we fully exploited the features of the dedicated DSP48E2 embedded in the FPGA and achieved the highest GOP/s compared with the
existing accelerator designs. To improve memory efficiency, we designed a synaptic weight delivery hierarchy and a Psum-Vmem unified buffer to support the high parallelism. To demonstrate FireFly's reconfigurability, we evaluated multiple deep SNN models on various datasets. To make SNN applications more convenient, we used off-the-shelf commercially available FPGA edge devices, offering a more feasible solution than any other existing hardware. In the future, we will try to migrate more optimization techniques targeting FPGAs while exploring sparsity acceleration to enable more energy-efficient SNN software and hardware co-design.

\bibliographystyle{IEEEtran}
\bibliography{myIEEE,reference}
\end{document}